\documentclass[12pt]{article}
\usepackage[pagenumbers]{cvpr} 

\usepackage{graphicx}
\usepackage{amsmath}
\usepackage{amssymb}
\usepackage{booktabs}
\usepackage{float}

\usepackage[pagebackref,breaklinks,colorlinks]{hyperref}

\usepackage[capitalize]{cleveref}
\crefname{section}{Sec.}{Secs.}
\Crefname{section}{Section}{Sections}
\Crefname{table}{Table}{Tables}
\crefname{table}{Tab.}{Tabs.}

\def\cvprPaperID{*****} 
\def\confName{TBD}
\def\confYear{2023}

\DeclareMathOperator{\argmin}{argmin}

\begin{document}

\title{How to Backpropagate through Hungarian in Your DETR?}

\author{Lingji Chen, Alok Sharma, Chinmay Shirore, Chengjie Zhang, Balarama Raju Buddharaju \\
Motional\\
Boston, MA, USA\\
{\tt\small motional.com}
}
\maketitle

\begin{abstract}
   The DEtection TRansformer (DETR) approach, which uses a transformer encoder-decoder architecture and a set-based global loss, has become a building block in many transformer based applications. However, as originally presented, the assignment cost and the global loss are not aligned, i.e., reducing the former is likely but not guaranteed to reduce the latter. And the issue of gradient is ignored when a combinatorial solver such as Hungarian is used. In this paper we show that the global loss can be expressed as the sum of an assignment-independent term, and an assignment-dependent term which can be used to define the assignment cost matrix. Recent results on generalized gradients of optimal assignment cost with respect to parameters of an assignment problem are then used to define generalized gradients of the loss with respect to network parameters, and backpropagation is carried out properly. Our experiments using the same loss weights show interesting convergence properties and a potential for further performance improvements.
\end{abstract}

\section{Introduction}
\label{sec:intro}
The seminal DEtection TRansformer (DETR) paper \cite{carion2020end} views object detection as a
direct set prediction problem. Its main ingredients are a set-based global loss that forces unique predictions via bipartite matching, and a transformer encoder-decoder architecture. DETR, together with its later variants such as Deformable DETR \cite{zhu2020deformable}, has become a building block for many transformer based approaches, \eg \cite{sun2020transtrack,zeng2021motr,misra2021end,meinhardt2022trackformer,ma2022unified,bai2022transfusion,zhou2022global,cai2022memot} in the area of detection and tracking, to name just a few.

In \cite{carion2020end}, the network produces a set of predictions whose number is larger than the number of ground truth boxes. An assignment problem is defined and solved by the Hungarian algorithm, and the optimally matched boxes are used to define a global loss to be minimized through backpropagation. The loss accounts for three aspects of the prediction: (1) The classes of the matched boxes should be those of their assigned ground truth, (2) the positions and sizes of these boxes should be their assigned ground truth, and (3) the classes of the non-matched boxes should be background. Intuitively, the criterion according to which the matching is made, \ie, the total assignment cost, should be aligned with the global loss, such that reducing the former necessarily reduces the latter.

But this is not the case in \cite{carion2020end}, because the matching cost is defined differently from the loss. The terms accounting for classes use raw probability instead of cross entropy; the heuristic reason given is relative scale to the loss from geometry. One may argue that the scaling problem is the same as in the loss, and can be explicitly dealt with using an additional scaling hyperparameter. The above terms are also limited to only the matched boxes; the reason given is not very convincing, because each prediction has a different probability of being background and therefore the missing sum is not  matching-independent. 

In a larger context, we may view the Hungarian solver as just another module performing some operations; in this case, some discrete optimization. One may argue that it does not matter how the cost matrix is defined, {\em as long as we can account for the gradient properly}. Unfortunately this is not done in most DETR approaches. In the code released with the paper \cite{carion2020end}, the issue of gradient is ignored by surrounding the {\tt matcher} code with {\tt torch.no\_grad()}, \ie, gradient tracing is turned off when Hungarian solver is involved.

In this paper we show that the matter is actually quite simple. First, the global loss can be expressed as a sum of two terms: One is defined by the probabilities of being background of {\em all predictions}, regardless of matching. The other can be treated as the optimal cost of matching, {\em if the cost matrix is suitably defined}.

Thus, to perform backpropagation on the global loss, we only need to answer the question: What is the gradient of the optimal cost with respect to the parameters defining the assignment problem? Fortunately this question, or rather a more general question involving Integer Linear Programming, has been answered in \cite{gao2019combinatorial}.

We have applied the new solution to the same training problem as in \cite{carion2020end}, and observed that the new solution produces desirable results, i.e., loss decreases, convergence shows, and the model performance improves as training progresses. Note that very limited experiments have been conducted so far and the authors will be performing more extensive experimentation in future work.

This paper is organized as follows. In \cref{sec:related} we introduce the reader to the larger problem of differentiability when the network contains an optimization module. In \cref{sec:main} we present our new solution that not only aligns the matching cost with the global loss, but also simplifies the implementation. Experiments for the same training problem as in \cite{carion2020end} with the proposed modifications are presented in \cref{sec:exp} to show the efficacy of the new algorithm, and \cref{sec:con} summarizes the key takeaways. We plan to release our code.

\section{Related Work}
\label{sec:related}
     \subsection{The general framework of Deep Declarative Networks} 
In \cite{gould2019deep} a new framework with end-to-end learnable models is presented, where a {\em declarative node} is defined by its desired behavior, in contrast to a conventional, {\em imperative node} that is defined by an explicit forward function. The desired behavior is typically stated as an objective and constraints in a mathematical optimization problem, and backpropagation can be performed, when the domain is continuous, with the help of the implicit function theorem, without having to know the actual solver used within. Previous work that inspired this generalization includes \cite{agrawal2019differentiable} that proposes an approach to differentiating through disciplined convex programs, and \cite{gould2016differentiating} that presents results on differentiating parameterized argmin and argmax problems in a lower-level which binds variables in the objective of an upper-level problem.

Since in a DETR based approach, a module solves an assignment problem, it can be considered a declarative node. However, because it involves discrete optimization, the gradient is not obtained through implicit function theorem, but through some other means, which will be described in the following sections.

\subsection{Differentiating combinatorial solvers by approximating the piecewise constant solution function}
An interesting, general approach is proposed in \cite{vlastelica2019differentiation} to approximate the piecewise constant solution function with an informative, differentiable function for a discrete optimization problem with a linear objective function. More specifically, to use the assignment problem as an example, let $x$ be the input data together with the network weight (at the current iteration), $I$ be the indices selected by the optimal solution, $C$ be the optimal weight, and $L$ be the loss to be minimized. If we start with $\partial L / \partial I$, then we would need $\partial I / \partial x$ to complete the backpropagation. However, the solution function from $x$ to $I$ is piecewise constant because there are only a finite number of values for $I$, and therefore its gradient is mostly zero or undefined. This is overcome in  \cite{vlastelica2019differentiation} by approximating the function with one that is differentiable and informative with non-zero gradients. 

We would argue that it is often more convenient to specify the loss in terms of the optimal value (rather than the optimal solution), and start with $\partial L / \partial C$. Thus the utility of \cite{vlastelica2019differentiation} is limited.

\subsection{Sinkhorn operator based approaches for matching}
To solve an assignment problem is to find a permutation matrix, where each row and each column has one and only one entry being ``1'' while the rest being ``0.'' This can be cast as a learning problem where the 1's are the truth labels and the network output is interpreted as the probabilities of the entries. However, these probabilities have a structural constraint, \ie, the matrix they form should be a doubly stochastic matrix (DSM) whose rows and columns all sum up to one. A bi-level optimization problem can be formulated where the lower level finds the best DSM approximation to the network output, which can be obtained through an iterative procedure called Sinkhorn-Knopp algorithm \cite{sinkhorn1967concerning}. The gradient can be computed by unrolling the iterations which are differentiable. This is essentially the approach in \cite{cruz2018visual,mena2018learning,papakis2020gcnnmatch} and many recent works.

Since unrolling the iterations takes a lot of memory, \cite{eisenberger2022unified} gives a closed form expression of the gradient in the form of solutions to linear equations, based on the Karush–Kuhn–Tucker
(KKT) conditions at the optimal value presumablly achieved at the end of the iterations.

\subsection{Differentiable assignment solver through relaxation and projection}
In \cite{zeng2019dmm} the Integer Linear Programming (ILP) assignment problem is relaxed to a Linear Programming (LP) problem, and Dykstra's algorithm \cite{dykstra1983algorithm} is used to project the solution into feasible space. This makes the solver differentiable.

\subsection{Network approximated assignment solvers}
To obtain differentiability, an approach is to approximate a combinatorial solver with a neural network whose operations are differentiable. This is achieved in \cite{lee2018deep,xu2020train,liu2022glan,psalta2022transformer} and others. In \cite{xu2020train}, non-differentiable tracking metrics are also approximated by neural networks.

\section{Our New Solution}
\label{sec:main}
\subsection{The global loss, rearranged}
We adopt some of the notations in \cite{carion2020end} but also define new ones. Let $y = \{y_j\}_{j=1}^M$ be the set of $M$ ground truth objects, and $\hat{y}=\{\hat{y}_i\}_{i=1}^N$ the set of $N$ predictions. At inference time, thresholding probabilities or scores will give us a subset of the $N$ predictions as actual objects. At training time, we want to select exactly $M$ out of $N$ predictions to correspond to the ground truth. Since typically $N \gg M$, we want to solve an assignment problem with a {\em rectangular} cost matrix (as is done in the {\em code} released by \cite{carion2020end}, not the {\em paper} itself), not a {\em square} one. Thus, instead of talking about a permutation, we define an injective mapping $s: G \to B$ from the ground truth index set $G \triangleq \{1, 2, \ldots, M\}$ to the predicted box index set $B \triangleq \{1, 2, \ldots, N\}$. Define the matched set $B_1 \triangleq s(G)$, then the unmatched set is $B_2 \triangleq B \backslash B_1$. The inverse mapping $s^{-1}: B_1 \to G$ is naturally defined. In other words, starting from the ground truth indexed by $j$, we find the assigned prediction indexed by $s(j)$. And starting from a prediction indexed by $i$ and already known to have had a ground truth assigned to it, we find the ground truth index $s^{-1}(i)$.  

As in \cite{carion2020end}, let $y_j = (c_j, b_j)$ be the ground truth with object class label $c_j$ and bounding box vector $b_j$. Let $\mathcal{L}_{\rm box}(b_j, \hat{b}_{s(j)})$ be the loss of predicted bounding box $\hat{b}_{s(j)}$ when it is considered to represent $b_j$. Let $\hat{p}_i\left(c_{s^{-1}(i)}\right)$ denote the probability of the target object class in the $i$-th prediction, and $\hat{p}_i(\emptyset)$, of background.

For a given set of network weight vector $w$, and a given mapping $s$, the loss has to do with three parts:
\begin{enumerate}
 \item the object class probabilities in the matched set $B_1$,
 \item the background probabilities in ``the rest'' set $B_2$, and 
 \item the losses $\mathcal{L}_{\rm box}$ between $G$ and $B_1$.
\end{enumerate}
The third part is the same as in \cite{carion2020end} and details will be omitted in this paper. The first two parts depend on the likelihood
\begin{equation} \label{eq:total_prob}
L \triangleq \prod_{j \in B_1} \hat{p}_j\left(c_{s^{-1}(j)}\right) 
 \prod_{k \in B_2} \hat{p}_k(\emptyset).
\end{equation}
Note that both terms depend on the mapping $s$. However, the following is a term that does not depend on the mapping $s$:
\begin{equation} \label{eq:const}
L_\emptyset \triangleq \prod_{i \in B} \hat{p}_i(\emptyset). 
\end{equation}
Thus the ratio
\begin{equation} \label{eq:ratio}
 \frac{L}{L_\emptyset} = \prod_{j \in B_1} \frac{\hat{p}_j\left(c_{s^{-1}(j)}\right)}{\hat{p}_j(\emptyset)}
\end{equation}
only has to do with the index sets $G$ and $B_1$.  The global loss (which is called Hungarian loss in \cite{carion2020end}) is therefore
\begin{equation} \label{eq:loss}
\begin{split}
 \mathcal{L}(y, \hat{y}, s) &= -\log \left(\frac{L}{L_\emptyset}\right) - \log (L_\emptyset) + \sum_{j=1}^M \mathcal{L}_{\rm box}(b_j, \hat{b}_{s(j)}) \\
&=  \sum_{j=1}^M \left( -\log \hat{p}_{s(j)}(c_j) + \log \hat{p}_{s(j)}(\emptyset) + \mathcal{L}_{\rm box}(b_j, \hat{b}_{s(j)}) \right) - \sum_{i=1}^N \log \hat{p}_i(\emptyset).
 \end{split}
\end{equation}

\subsection{The new cost matrix}
In the above loss, we will try to write the matching dependent terms as a total assignment cost, \ie,
\begin{equation} \label{eq:loss2}
\begin{split}
 \mathcal{L}(y, \hat{y}, s) &= \sum_{j=1}^M \left( -\log \hat{p}_{s(j)}(c_j) + \log \hat{p}_{s(j)}(\emptyset) + \mathcal{L}_{\rm box}(b_j, \hat{b}_{s(j)}) \right) - \sum_{i=1}^N \log \hat{p}_i(\emptyset) \\
&\triangleq \sum_{j=1}^M \mathcal{C}(s(j), j) + \mathcal{L}_{\rm background},
 \end{split}
\end{equation}
where we define the $(i, j)$-th entry of an $N \times M$ cost matrix $\mathcal{C}$ as 
\begin{equation} \label{eq:costmat}
 \mathcal{C}(i, j) \triangleq -\log \hat{p}_{i}(c_j) + \log \hat{p}_{i}(\emptyset) + \mathcal{L}_{\rm box}(b_j, \hat{b}_i).
\end{equation}
It is now clear that to minimize the global loss, we take two steps:
\begin{enumerate}
 \item Solve an assignment problem using the cost matrix defined in \cref{eq:costmat}:
 \begin{equation} \label{eq:s}
  \hat{s} = \underset{s: G \to B}{\argmin} \sum_{j=1}^M \mathcal{C}(s(j), j).
 \end{equation}
\item Minimize 
\begin{equation} \label{eq:L}
\mathcal{L}(y, \hat{y}, \hat{s}) \triangleq \mathcal{L}_{\rm assign} + \mathcal{L}_{\rm background}.
\end{equation}
\end{enumerate}

\subsection{The generalized gradients}
The generalized gradients of the optimal assignment cost $\mathcal{L}_{\rm assign}$ with respect to the network weights $w$ can be obtained using the solution given in \cite{gao2019combinatorial}. Thus the generalized gradients of $\mathcal{L}(y, \hat{y}, \hat{s})$ can be obtained and used in training \cite{clarke1990optimization}.

More specifically, the cost matrix $\mathcal{C}$ in \cref{eq:costmat} is defined by the ground truth $y$ and the network predictions $\hat{y}(w)$ as a function of the network weight vector $w$. If we stack up the columns of $\mathcal{C}$ to form a cost vector $c(w)$, then the assignment problem can be formulated as an Integer Linear Programming (ILP) problem with parameters $(c(w), A, b)$ where both $A$ and $b$ are constants specifying the inequality constraints of a valid assignment \cite{crouse2016implementing}. -- We can pad the cost matrix with almost zero random values to form a square one with equality constraints, so that the results in \cite{gao2019combinatorial} can be applied.

Let $\hat{u}$ be an optimal solution (in the form of a column vector) to the assignment problem; practically speaking it is almost always unique. Then according to Algorithm 1 in \cite{gao2019combinatorial}, we have
\begin{equation} \label{eq:grad}
 \frac{\partial \mathcal{L}_{\rm assign}}{\partial w} = \hat{u}^T \frac{\partial c}{\partial w}.
\end{equation}

This gives us a way to calculate the gradients with {\tt pytoch}: We treat the optimal assignment solution $\hat{u}$ as a constant; this means that we take only the {\em numerical values} of the cost matrix and call a Hungarian solver, under {\tt torch.no\_grad()}. Then we can sum up the gradient-attached optimal cost according to the solution $\hat{u}$, together with the background terms, and then backpropagate the loss in \cref{eq:L}.

\subsection{The differences}
We belabor the point by noting the following differences of our new solution compared to \cite{carion2020end}:
\begin{itemize}
 \item The assignment cost and the global loss are now aligned, and this is conceptually satisfying.
 \item The unmatched predictions have different probabilities of background in them that depend on the assignment, and therefore we do not ignore them in the cost matrix.
 \item The cross entropy loss is used in our cost matrix as opposed to the raw probabilities; the problem with scaling, if it still exists, can be solved separately.
 \item The challenge of the rectangular cost matrix is solved by using the ratio in \cref{eq:ratio} that results in the cost matrix in \cref{eq:costmat}, \ie, every row subtracts the cross entropy loss that corresponds to the background.
 \item There is a normalization by the number of boxes in their global loss definition for GIOU based losses, which is not done in their total assignment cost. The loss as expressed by \cref{eq:loss} does not justify the normalization. After all, a batch having more boxes {\em should} have a bigger impact.
\end{itemize}

\section{Experiments}
\label{sec:exp}
We perform quantitative evaluation of the proposed global loss and cost matrix formulation using DETR architecture\cite{carion2020end} on the COCO dataset\cite{coco2014dataset}. For a comparison of performance, we first train DETR as is, and then train the model again using the proposed Assignment Aligned DETR as described in the previous section.

\subsection{Dataset and Metrics}We use COCO2017 detection dataset\cite{coco2014dataset}, containing 118k training and 5k validation images. The images are annotated with bounding boxes and segmentation but for our purposes only the bounding boxes are relevant.
For comparison with the DETR methodology, we report the integral AP metric used by  COCO. This AP is averaged over multiple Intersection over Union (IoU) values. We also present plots showing how  COCO AP metrics evolve over training epochs. 

\subsection{Training Details}For our experiments, we train DETR using the source code provided by its authors on \href{https://github.com/facebookresearch/detr}{github}. For the baseline using bi-partite matching loss as well as the proposed global loss, the model is trained as is using DETR setup hyper-parameters  except that we use a 100 epoch training schedule, batch size of 2 per GPU and with auxiliary losses disabled. Note that DETR was trained using a 300 epoch training schedule with learning rate drops.

\subsection{Experiment Results}
\begin{figure}[H]
\begin{subfigure}{0.5\textwidth}
    \centering
    \includegraphics[scale=0.55]{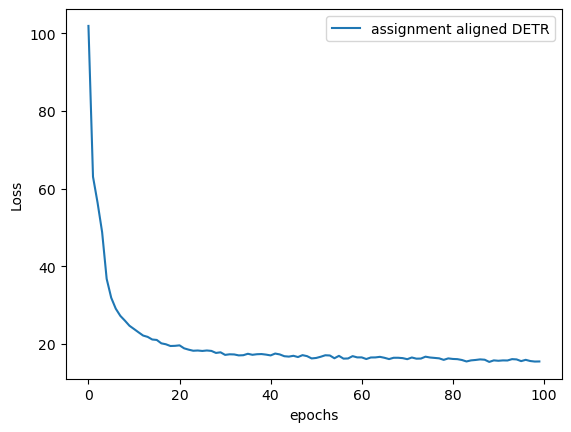}
    \caption{Loss Convergence over epochs}
    \label{fig:loss_convergence}
\end{subfigure}%
\begin{subfigure}{0.5\textwidth}
  \centering
  \includegraphics[scale=0.55]{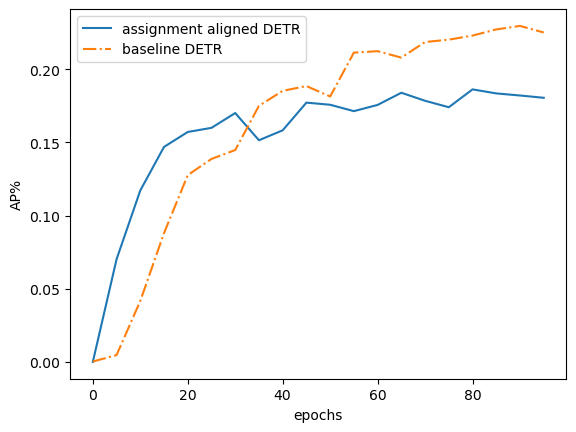}
  \caption{AP\ vs\ Epoch}
  \label{fig:COCO AP}
\end{subfigure}
\begin{subfigure}{.5\textwidth}
  \centering
  \includegraphics[width=\linewidth]{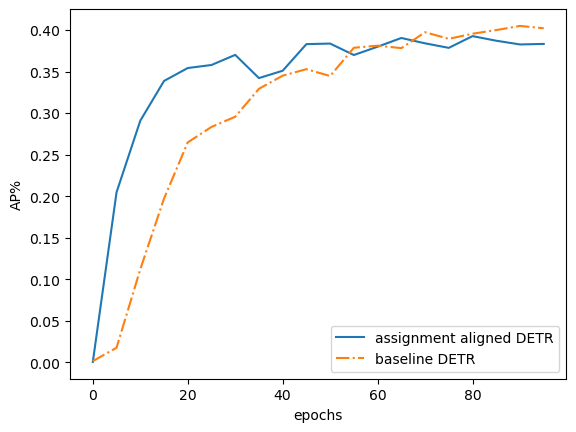}
  \caption{AP\textsubscript{50}\ vs\ Epoch}
  \label{fig:AP05}
\end{subfigure}%
\begin{subfigure}{.5\textwidth}
  \centering
  \includegraphics[width=\linewidth]{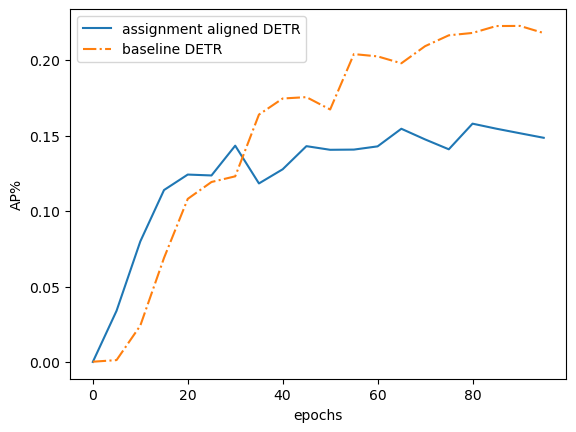}
  \caption{AP\textsubscript{75}\ vs\ Epoch}
  \label{fig:AP075}
\end{subfigure}
\caption{Results of training DETR with proposed loss for 100 epoch cycle}
\label{fig:AP_plots}
\end{figure}

Results of evaluation on COCO2017 validation set after 100 epochs are presented in  Table \ref{table:1}. We also present plots showing how loss values changed over the 100 epoch training cycles in Fig \ref{fig:loss_convergence} along with plots showing how the AP changed during the experiment. 

The loss plots in Fig\ref{fig:loss_convergence} indicate that with the new formulation, the loss consistently decreased and exhibited signs of convergence. The AP plots in Fig 
 \ref{fig:COCO AP}, Fig \ref{fig:AP05} \& Fig \ref{fig:AP075} also show that the network is consistently able to learn the classes from data and its performance increased as the training progressed. If compared to the baseline DETR, the plots indicate that with the new loss formulation, the network performance did not exceed the baseline in the limited amount of experiments performed. 
 However, it should be noted that the overall trend of metrics shows that conceptually speaking, the proposed mathematical formulation where the matching cost aligns with the global loss, produces desirable results specifically as loss decreases the network performance increases.

\begin{table}[h!]
\centering
\begin{tabular}{c  c  c c  c  c  c  c  c} 
 \hline
 Model & Matching Strategy & Loss Minimized & AP & AP\textsubscript{50} & AP\textsubscript{75} & AP\textsubscript{S} & AP\textsubscript{M} & AP\textsubscript{L}\\ [0.5ex] 
 \hline
 DETR -R50 & Hungarian  & Set-Based Loss & 24.1 & 42.40 & 235.5 & 8.5 & 25.1 & 39.0 \\ 
 \hline
 DETR -R50 & Hungarian  & Assignment Cost & 18.90 & 39.5 & 15.9 & 4.10 & 18.7 & 34.9 \\
 \hline
\end{tabular}
\caption{Table shows the results of evaluation of DETR trained using loss defined in \cite{carion2020end} and proposed loss implementation on COCO validation set .}
\label{table:1}
\end{table}

\section{Conclusion}
We have presented a new formulation of assignment cost that can be applied to DETR like architectures which use Hungarian matching. This cost is aligned with the global loss calculation, and leads to proper use of the gradient even with the presence of a Hungarian solver. With the limited experiments conducted using DETR code-base, it can be concluded that the formulation is valid and produces desirable results. 
Given the promising results with initial experiments, the authors expect that matching SOTA results with a hyper-parameter search will be feasible, and such experiments will be part of the authors' future work. The authors also expect that the theoretically sound approach will encourage the research community to develop further and produce better performing models.

\label{sec:con}


\nocite{*}

{\small
\bibliographystyle{ieee_fullname}
\bibliography{cvpr23}
}

\end{document}